# OCFER-Net: Recognizing Facial Expression in Online Learning System*

Yi Huo[†]
AI& Edu Lab, Teacher's College
Beijing Union University
Beijing China
sfthuoyi@buu.edu.cn

Lei Zhang
College of Computer Science
Comm. University of China
Beijing State China
leizhang@cuc.edu.cn

## ABSTRACT

Recently, online learning is very popular, especially under the global epidemic of COVID-19. Besides knowledge distribution, emotion interaction is also very important. It can be obtained by employing Facial Expression Recognition (FER). Since the FER accuracy is substantial in assisting teachers to acquire the emotional situation, the project explores a series of FER methods and finds that few works engage in exploiting the orthogonality of convolutional matrix. Therefore, it enforces orthogonality on kernels by a regularizer, which extracts features with more diversity and expressiveness, and delivers OCFER-Net. Experiments are carried out on FER-2013, which is a challenging dataset. Results show superior performance over baselines by 1.087. The code of the research project is publicly available on https://github.com/YeeHoran/OCFERNet.

## 1 Introduction

With the rapid development of artificial intelligence and big data, online learning appears by integrating them with traditional education and is very popular. It provides ubiquitous learning so that anyone, at any time, in anywhere, can learn anything online with any terminal devices(5A), which promotes knowledge dissemination, distribute high-quality educational resources more efficiently, and contributes to education equity. The entry of COVID-19 has become a catalyst for online learning, makes it develop rapidly and become normal. The intelligent requirements for online learning have also increased. Students not only want to obtain knowledge, but also want to improve emotion interaction with it. They hope to have the same emotional interaction experiences as in real classroom environments.

In traditional classroom, students' cognitive change will trigger various positive or negative emotions, which are shown in students' facial expressions. Experienced teachers can infer their cognition level about the knowledge by observing their expressions and adjust teaching strategies accordingly. However, in online learning platform, especially in large classes with hundreds of students, it is difficult for teachers to see the facial expressions of all students online in real time. Even though the two-way interactive video interaction can start, the data amount of all students' videos to be transmitted and processed would be extremely huge. Therefore, it is difficult for teachers to obtain students' cognition situation from this aspect.

Therefore, it adds intelligent FER tool to online learning system in the project. Through analyzing students' facial expression features in online learning environment, the facial expression recognition system detects students' emotion category in real time. After obtaining the emotional categories of all students, teachers can adjust instruction accordingly. Therefore, the performance of facial expression recognition (FER) system affects the online learning efficiency in a large extent and has a direct impact on the sustainable development of intelligent online learning system.

Publicly known basic emotions are anger, disgust, fear, happiness, sadness, surprise and neutral. Many researchers tried to improve the accuracy of FER. Reference[1][2] found in natural environment, due to occlusion, posture change, illumination change and other reasons, the difference of intra-class becomes large, while the difference of inter-class decreases. Yumin Tian[1] add two fully-connected layers of a novel quadruplet-mean loss on traditional deep convolutional neural network to enlarge intra-class similarity and inter-class distinction. Yangtao Du[2] introduces an Expression Associative Network to learn association of facial expression and auxiliary module as the invariant feature generator on Generative Adversarial Networks to suppress pose variations, illumination changes, and occlusions. To enhance generalization of extracted features in FER, Liam Schoneveld[3] trains them on multiple FER datasets, and employs knowledge distillation, alongside additional unlabeled data. Farzaneh[4] dicovers that equally supervising all features might include irrelevant features and degrade the generalization ability, and proposes a Deep Attentive Center Loss method to adaptively select a subset of significant features to enhance discrimination. In fact, feature redundancy reduces FER performance as well. Nevertheless, current work ignores cutting down feature redundancy towards performance improvement in FER. Di Xie[5] points that regularizer that utilizes orthonormality among different filters can alleviate the problem. To enforce orthonormality on convolutional filters, Jiayun Wang[6] further introduces orthogonal convolutional matrix which presents good performance in classification problems.

These works are important to improve FER accuracy across various applications and is inspired by their ideas, makes full use of the orthogonality of convolution matrix, extracts more diverse and expressive features and based on which the OCFER-Net is proposed. Experiment shows that this method can effectively improve the FER performance. The implementation code is publicly available and the proposed FER method can be verified.

## 2 The Proposed OCFER-Net

In this section, it first presents the architecture of OCFER-Net, and then introduce the proposed orthogonal convolutional loss function which intends to achieve more diverse and expressive features. Finally, the overall objective function of OCFER-Net will be introduced.

### 2.1 OCFER-Net

The network architecture is based on ResNet-18. The orthogonal convolutional regularizer will select some layers from the network to impose orthogonality loss on them.

### 2.2 Orthogonal Convolutional Loss Function

The orthogonality loss is developed by kernel orthogonality condition [6]. For a convolutional layer with input tensor $X \in R^{C \times H \times W}$ kernel $K \in R^{M \times C \times k \times k}$, and the output tensor $Y = Conv(X, K)$, then the Orthogonal convolution condition is show in (1), where $I_{r0} \in R^{M \times M \times (\frac{2P}{S}+1) \times (\frac{2P}{S}+1)}$ is a tensor with zero entries except the enter $M \times M$ entries are identity matrix. Equation (2) gives a near-orthogonal convolution regularization.

$$Conv(K, K, padding = P, stride = S) = I_{r0} \qquad (1)$$

$$\min_K \|Conv(K, K, padding = P, stride = S) - I_{r0}\|_F^2 \qquad (2)$$

### 2.3 Overall Objective Loss Function

The overall objective function is obtained by combining the final loss and orthogonal convolution regularization loss, so that the FER objective task and feature diversity could be achieved at the same time. Denoting $\lambda > 0$ as the weight of orthogonal regularization loss, then the overall loss function is (3), where $L_{task}$ is the FER task loss, and $L_{orth}$ is the orthogonal regularization loss.

$$L = L_{task} + \lambda L_{orth} \qquad (3)$$

## 3 Experiment

The proposed OCFER-Net is compared with the most popular FER system in GitHub [7] on the challenging FER2013 Dataset. It is a large-scale dataset collected by Google image search which contains 28,709 training images, 3,589 validation images and 3,589 test images.

### 3.1 Implementation Details

The experiment uses training set to train OCFER-Net, 3589 images for public test, and 3,589 images for private test. The DFER-Net is trained by cross entropy loss function and stochastic gradient descent with batch size of 8. The learning rate is set to 0.01 initially and is reduced by a factor of 10 every 10,000 iterations. The total epochs are 250 and is set to 0.5.

### 3.2 Experimental Results

The comparison results of different methods are reported in Table I. The proposed OCFER-Net has a trivial decline of 0.362% in public test, but with a substantial improvement of 1.087% over baselines[7] on FER2013. OCFER-Net forces orthogonal convolution and thus learn more diverse features. This shows the effectiveness of the proposed orthogonal regularization constraint.

Table 1: Recognition performance of different methods

| Methods | Public test(%) | Private test(%) |
|---|---|---|
| WuJie1010[7] | 71.580 | 72.388 |
| Proposed OCFER-Net | 71.218 | 73.475 |

The confusion matrices of OCFER-Net on FER2013 is reported in Table 2. It shows the detailed classification result for each expression. From Table II, it can be observed that OCFER-Net achieves better accuracy in happy, surprise and neutral expressions than angry, disgust, fear and sad expressions. This is acceptable since it is even difficult for person to recognize these expressions.

Table 2: Recognition performance of different methods

|    | AN | DI | FE | HA | SA | SU | NE |
|----|----|----|----|----|----|----|----|
| AN | 0.68 | 0.01 | 0.08 | 0.02 | 0.11 | 0.01 | 0.08 |
| DI | 0.18 | 0.71 | 0.05 | 0.04 | 0 | 0 | 0.08 |
| FE | 0.09 | 0 | 0.56 | 0.02 | 0.18 | 0.07 | 0.08 |
| HA | 0.01 | 0 | 0.01 | 0.9 | 0.03 | 0.01 | 0.03 |
| SA | 0.08 | 0 | 0.1 | 0.03 | 0.63 | 0.01 | 0.15 |
| SU | 0.03 | 0 | 0.08 | 0.03 | 0.02 | 0.83 | 0.01 |
| NE | 0.05 | 0 | 0.05 | 0.03 | 0.12 | 0.01 | 0.73 |

## 4 Conclusion

In this paper, the OCFER-Net is proposed to recognize facial expression images. It improves the FER performance through extracting more diverse and expressive features. In addition, it enforces orthogonal convolution on filters by a regularizer. Experimental results verify the superiority of OCFER-Net in facial expression recognition.

## ACKNOWLEDGMENTS

Supported by the Academic Research Projects of Beijing Union University (No. ZK30202110).